\documentclass[conference]{IEEEtran}
\IEEEoverridecommandlockouts

\usepackage{eso-pic}
\usepackage{cite}

\usepackage{amsmath,amssymb,amsfonts}
\usepackage{algorithmic}
\usepackage{graphicx}
\usepackage{textcomp}
\usepackage{xcolor}
\usepackage{booktabs}
\usepackage{multirow}
\usepackage{url}
\def\BibTeX{{\rm B\kern-.05em{\sc i\kern-.025em b}\kern-.08em
    T\kern-.1667em\lower.7ex\hbox{E}\kern-.125emX}}
\begin{document}

\bstctlcite{IEEEexample:BSTcontrol}
\newcommand{\Houbo}[1]{\textcolor{purple}{#1}}
\newcommand{\Shikai}[2]{\textcolor{red}{#2}}

\title{RF-Agent: A Practical Framework for Building Language Agents for RFIC Design
}

\author{
    Yueqi Xing\textsuperscript{1*}, 
    Houbo He\textsuperscript{1*}, 
    Jolie Wang\textsuperscript{1}, 
    Erin Ni\textsuperscript{1},
    Shikai Wang\textsuperscript{2}, 
    Qiufeng Li\textsuperscript{2}, 
    Weidong Cao\textsuperscript{2}, 
    Taiyun Chi\textsuperscript{1} \\
    \textsuperscript{*}Equally Credited Authors (ECAs) \quad
    \textsuperscript{1}ECE Department, Rice University, Houston, TX, USA \\
    \textsuperscript{2}ECE Department, The George Washington University, Washington, D.C., USA \\
    \{xy67, hh68, jw161, ejn5, taiyun.chi\}@rice.edu, \{shikai.wang, qiufeng.li, weidong.cao\}@gwu.edu
}



\maketitle
\AddToShipoutPictureFG*{%
  \AtTextLowerLeft{\raisebox{-18pt}{\footnotesize 979-8-3195-1246-8-0/26/\$31.00~\copyright2026~IEEE}}%
}
\begin{abstract}
Large language models (LLMs) have driven rapid progress in electronic
design automation (EDA), yet their application to radio-frequency (RF)
circuit design remains limited by the scarcity of domain-specific
datasets and standardized benchmarks. We present RF-Agent, which
addresses this gap through textbook-driven knowledge distillation. A
multi-agent Question--Thinking--Solution--Answer (QTSA) pipeline
converts a subsection-level corpus from seven canonical RF textbooks
into the first-of-its-kind RF-domain reasoning dataset (over 11,000
samples) with a dedicated multiple-choice benchmark. On this benchmark
we study two adaptation strategies: supervised fine-tuning (SFT) and
three retrieval-augmented generation (RAG) configurations (semantic,
keyword, hybrid). Across multiple LLM families, domain-specific SFT
significantly improves RF reasoning, especially for small and
medium-sized models; among RAG configurations, semantic retrieval
performs best, indicating embedding-based context alignment suits RF
reasoning better than naive fusion. The dataset and benchmark provide a
reusable foundation for future work on LLM-aided RF circuit design.
\end{abstract}

\section{Introduction}
 
Large language models (LLMs) have demonstrated remarkable capabilities in natural language understanding, code generation, and scientific reasoning, and their adoption in EDA has grown rapidly. Recent work spans analog circuit synthesis and topology design~\cite{analoggenie,lamagic,analogcoder,analogxpert,WeidongICCAD}, parameter optimization~\cite{ADOLLM,ledro,analog_size,toposizing}, and RTL generation for digital design~\cite{rtlcoder,verigen,chipnemo}. However, recent evaluations show that general-purpose LLMs still exhibit significant deficiencies in circuit reasoning for both analog~\cite{RazaviQA,RazaviQA2} and digital circuits~\cite{verilogeval,chipnemo}, with these limitations growing
more severe as circuit complexity and domain specialization increase. A growing body of work addresses this through \textit{domain-adapted QA and reasoning
agents} that ground model responses in circuit knowledge via domain-specific fine-tuning
and RAG~\cite{ampagent,askeda,MuaLLM,chipnemo,
analogseeker,eda_raft,AMS_KG}.

\textbf{The RF-specific gap.} RF circuits form the backbone of
modern wireless systems, from 5G transceivers to
radars and satellite links~\cite{fem_isscc,array_isscc,dband_cicc}, making their efficient design critical
to continued advances in connectivity and sensing~\cite{xiaohan_PA,hao_vco,MWSCAS}. Despite progress in analog and digital EDA, and despite machine-learning methods advancing inverse design of RF building blocks~\cite{eth_review,KaushikNC,Chae_ASPDAC_2025,motifrf, adreamdco}, RF domain remains substantially underexplored for LLM-based
design assistance~\cite{chae2025survey}. Three factors contribute to this gap. First, RF circuit knowledge is concentrated in specialized textbooks and proprietary design notes, with no curated open repositories. Second, RF reasoning requires
structured multi-step derivation, spanning topology recognition, S-parameter analysis, impedance matching, and more, which requires domain-specific analytical skills beyond what general pretraining offers~\cite{RazaviQA}. Third, while recent benchmarks have been released for analog circuits evaluation, no standardized RF benchmark exists, and no prior work has
simultaneously addressed RF-domain dataset and
benchmark development, and evaluation of domain-adaptation strategies.

In this work, we present RF-Agent, a framework that addresses
this gap through two complementary strategies: (1) constructing the first
RF-domain reasoning dataset and benchmark via a multi-agent QTSA (Question–Thinking–Solution–Answer) distillation pipeline, and (2) developing and evaluating RAG systems for RF-specific knowledge grounding.
Our dataset and code are publicly available at
\url{https://github.com/Nina-nina123/RF-Agent}.
Our main contributions are:

\begin{itemize}
    \item \textbf{Open RF reasoning dataset and benchmark}: The first
    RF-domain QTSA dataset with five-perspective question diversity
    and dual multiple-choice (mcQTSA) / normal dialog (ndQTSA) formats, yielding over 11,000 reasoning samples. The mcQTSA subset forms 
    the standardized RF multiple-choice
    benchmark, providing an objective evaluation resource previously
    absent from RF domain.
    \item \textbf{RF-specific retrieval augmentation:} A systematic
    comparative evaluation of three retrieval configurations, semantic embedding-based, keyword-based, and hybrid, for
    RF-domain knowledge grounding, conducted on a corpus of 950
    peer-reviewed papers and 7 canonical RF textbooks across
    multiple model families.
    \item \textbf{Cross-model domain adaptation analysis}: A
    systematic study across multiple LLM families (0.6B--4B) and
    state-of-the-art models including GPT, DeepSeek and Qwen,
    demonstrating that targeted domain adaption consistently
    improves RF reasoning and enables small fine-tuned models to
    approach the performance of large models.
\end{itemize}


\section{RF-Domain Dataset and Benchmark} \label{sec:dataset}

  \begin{figure*}[!t]
\centering
\includegraphics[height=6cm, trim=19 15 19 15, clip]{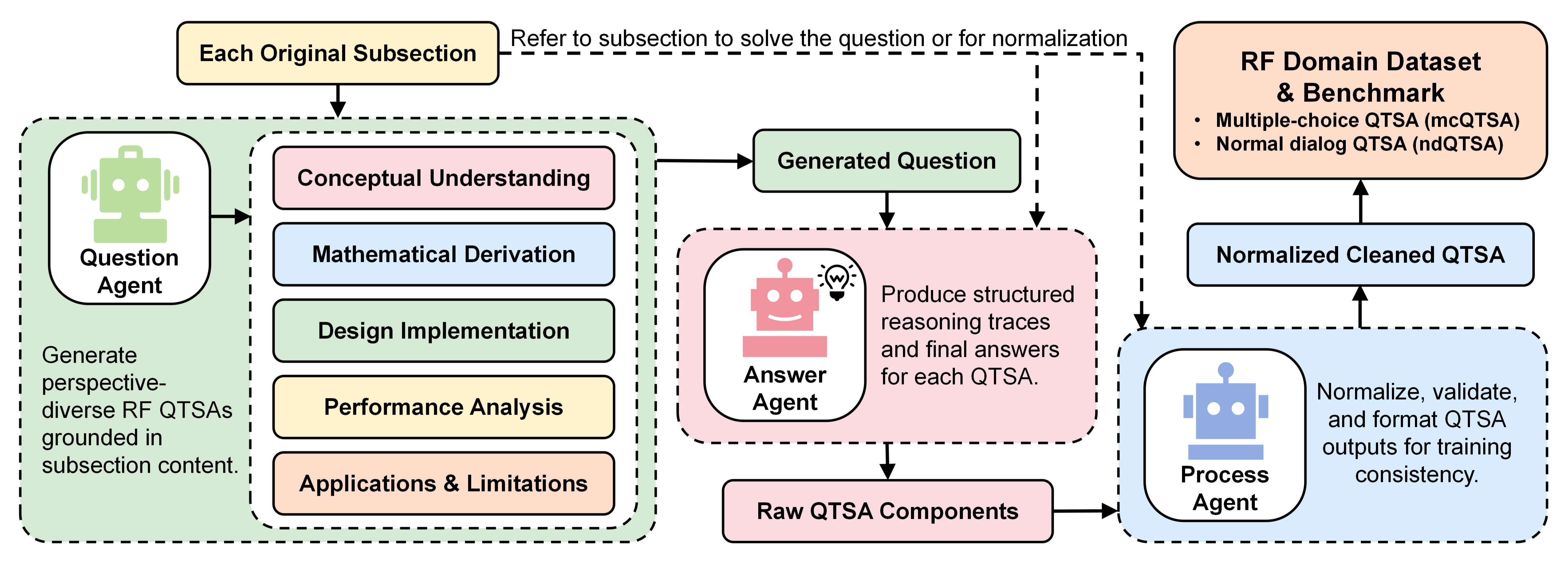}
\vspace{-0.8em}
\caption{Multi-agent QTSA distillation pipeline. The Question Agent generates perspective-diverse mcQTSA and ndQTSA questions from each original subsection. The Answer Agent produces structured CoT traces. The Process Agent normalizes and validates all outputs.}

\label{fig:qtsa_workflow}
\vspace{-1.8em}
\end{figure*}
 

Effective domain adaptation requires structured supervision that
captures multi-step reasoning, not simply raw text corpora. Prior work has shown that chain-of-thought (CoT) distillation from authoritative sources
transfers reasoning capability more effectively than answer-only
supervision~\cite{ho2023cot_teachers,feng2024kpod}. Multiple-choice benchmarks from expert-curated content further provide objective, automatically verifiable evaluation at low annotation cost, a critical advantage in domains where expert
labeling is scarce~\cite{amsbench,MMCircuitEval}.
 
Building on these insights, we construct the RF-specific reasoning dataset and benchmark using a multi-agent QTSA distillation pipeline that produces two complementary formats. \textbf{mcQTSA} pairs each question with four answer options and a full Question--Thinking--Solution--Answer quadruple, making it suitable for both SFT training and objective evaluation. \textbf{ndQTSA} generates open-ended question-answer pairs for explanatory reasoning. Our framework introduces three targeted extensions over prior QTSA work~\cite{analogseeker}: (1)~\textit{five-perspective} question generation to maximize coverage across reasoning types; (2)~a reasoning-oriented answer model that produces richer CoT traces for thinking-mode SFT; and (3)~the dual mcQTSA/ndQTSA design that serves training and evaluation within a single pipeline. Fig.~\ref{fig:qtsa_workflow} illustrates the overall pipeline.
 
\subsection{RF Corpus Construction}
The quality of training data matters as much as its quantity: prior work has shown that training on textbook-quality content enables small models to achieve strong performance compared to models trained on larger but noisier corpora~\cite{gunasekar2023textbooks}. Motivated by this finding, we ground our dataset construction in seven canonical RF textbooks spanning foundational theory, amplifier design, system analysis, and transceiver architecture, topics that are largely absent from prior analog-focused corpora and concentrated in expert-authored texts. Each book is parsed at the subsection level into self-contained conceptual units, yielding a corpus of 1,108 subsections.
 
\subsection{Multi-Agent QTSA Distillation} \label{sec:distillation}
 
We adopt a three-agent pipeline that converts each subsection into QTSA quadruples. This pipeline is executed five independent times per subsection, with each execution targeting a distinct reasoning perspective, to encourage diversity while remaining grounded in the source content.
 
\subsubsection{Question Agent} The Question Agent (GPT-4.1-mini) generates mcQTSA and ndQTSA questions strictly grounded in the provided subsection. 
For mcQTSA, each question includes one correct answer and three technically plausible distractors drawn from the same subsection, targeting understanding rather than surface recall. To promote reasoning diversity, five independent executions are performed under distinct reasoning perspectives, ensuring that repeated runs over the same subsection produce meaningfully different questions rather than paraphrases of one another. It is motivated by findings that reasoning-type diversity in distilled data significantly improves downstream performance~\cite{feng2024kpod}. To confirm this, we manually reviewed generated question sets across multiple subsections and verified that questions produced under distinct dimensions are meaningfully different from one another.
 
\subsubsection{Answer Agent} The Answer Agent (GPT-5-mini) generates the Thinking, Solution, and Answer fields using a reasoning-oriented model. Given the multi-step analytical nature of RF problems, using a reasoning model rather than a standard instruction model produces substantially richer CoT supervision~\cite{ho2023cot_teachers}.
 
\subsubsection{Process Agent} A deterministic agent (GPT-4.1-mini) enforces JSON compliance, validates field completeness, and strips corpus-specific references (equation numbers, figure indices), ensuring the generated samples are self-contained and distributable independently of the sources.
 
\subsection{Dataset Composition and Value}
 
The pipeline yields over 11,000 QTSA samples for multiple-choice and open-ended reasoning. As shown later with experiments, mixed-format training outperforms either format alone. 

The QTSA format supervises full reasoning trajectory rather than the final answer alone, aligning training directly with the inference behavior of thinking-mode models. The mcQTSA format further serves as a reusable RF benchmark, providing objective multiple-choice evaluation grounded in expert RF knowledge. This fills a critical gap, where no standardized RF LLM benchmark previously existed~\cite{amsbench,MMCircuitEval}.

\section{Methods} \label{sec:methods}

To adapt general-purpose language models to RF circuit reasoning, we pursue two complementary strategies: SFT on the QTSA dataset constructed in Section~\ref{sec:dataset}, and RAG grounded in an authoritative RF knowledge base. 
SFT encodes domain knowledge directly into model parameters, improving the model's ability for RF-specific analytical reasoning without external support at inference time. RAG, by contrast, decouples reasoning from parametric memory, allowing
the model to reference verified domain sources at inference time without
retraining, and unlike SFT applies even to API-only foundation models. Together, they provide a flexible framework that can be applied to models of varying sizes and capability levels.

\subsection{Supervised Fine-Tuning (SFT) on QTSA Data}

\subsubsection{Training Data Format}
We train directly on the QTSA samples from Section~\ref{sec:dataset}.
Reasoning-capable models (e.g., Qwen3 in thinking mode) learn the full
Q+T+SA format, with the thinking trace $T$ as a separate generation
phase; standard instruction-tuned models learn a Q+TSA format that
concatenates thinking, solution, and answer into a single sequence,
teaching non-reasoning models explicit reasoning patterns. The full
training schema is provided in the repository.
\subsubsection{Model Families and Configurations}
We evaluate SFT across two model families, LLaMA-3.2~\cite{llama} and
Qwen3~\cite{qwen}, spanning 0.6B--4B parameters and including both base
and instruction-tuned variants, enabling a systematic study of how
model capacity and prior alignment interact with domain-specific
supervision. For Qwen3, we evaluate both thinking-enabled and
thinking-disabled inference: the former generates an explicit CoT trace
before the answer, mirroring the QTSA training format, while the latter
produces a direct response, trading accuracy for inference efficiency.
\subsection{Retrieval-Augmented Generation over RF Knowledge Base}



\subsubsection{Knowledge Base Construction}
Our retrieval corpus consists of 950 RF-related peer-reviewed papers
and 7 canonical RF textbooks, forming a dense RF knowledge base. All
documents are split into overlapping text chunks and embedded with
BGE-M3~\cite{bgem3} into a ChromaDB vector database. Reflecting the
multimodal nature of RF literature~\cite{Mediatek}, diagrams
(schematics, Smith charts, S-parameter plots) are additionally
extracted via GPT-4o into textual descriptions and indexed in the
same store. Details are in the repository.




\subsubsection{Retrieval Pipeline}
\begin{figure*}[t]
    \centering
    \includegraphics[height = 4.8cm, trim=135 77 93 150, clip]{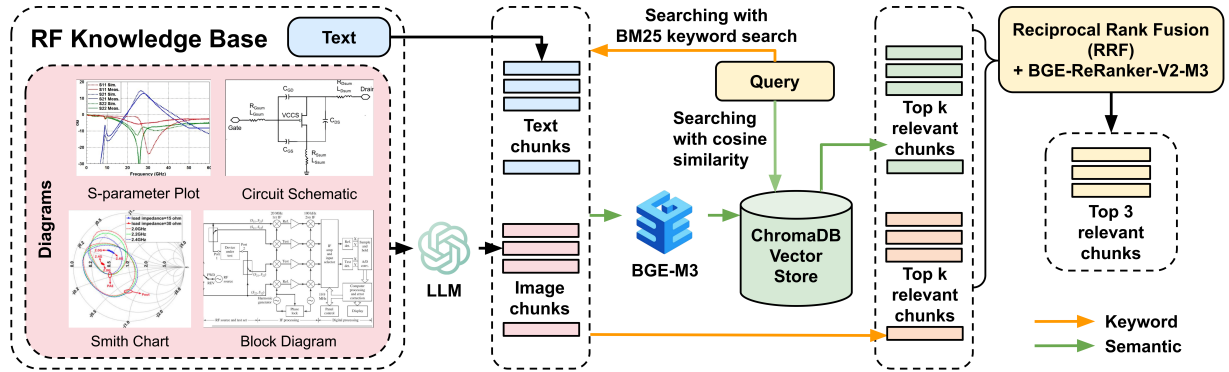}
    \vspace{-1em}
    \caption{Hybrid RAG pipeline for RF knowledge retrieval.}
    \label{fig:rag_pipeline}
\vspace{-1.5em}
\end{figure*}

We implement three retrieval configurations: two single-method
baselines, \textbf{semantic} and \textbf{keyword} retrieval, and a
\textbf{hybrid} configuration that combines both. The semantic
configuration embeds the query with BGE-M3 and retrieves the top-3
chunks by cosine similarity. The keyword configuration retrieves the
top-3 chunks by BM25~\cite{BM25} lexical overlap, capturing exact
component names, topologies, and abbreviations that dense embeddings
underweight in technical domains~\cite{askeda, MuaLLM}. The hybrid
configuration (Fig.~\ref{fig:rag_pipeline}) retrieves the top-15
candidates from each, merges them via Reciprocal Rank Fusion
(RRF)~\cite{rrf}, re-ranks with BGE-ReRanker-V2-M3~\cite{bgem3}, and
passes the top-3 chunks as context. The retrieved chunks
are prepended to the model's input prompt.




\vspace{-1em}
\section{Experiments}\label{sec:experiments}
All models are evaluated on a fixed 1,000-sample mcQTSA benchmark with
deterministic greedy decoding; an answer extractor maps the
outputs to a choice in \{A, B, C, D\}.
\vspace{-1em}

\subsection{Supervised Fine-Tuning Across Model Series}

We evaluate the effectiveness of the proposed QTSA dataset through
supervised fine-tuning across multiple model families.

\subsubsection{mcQTSA and ndQTSA Provide Complementary Supervision}
To investigate the effect of dataset format, we fine-tune
Llama3.2-1B-Instruct using three 5M-token subsets: mcQTSA, ndQTSA,
and a randomly mixed dataset. Training on mcQTSA alone achieves
54.9\%, outperforming ndQTSA-only training at 45.9\%, as it directly
trains the model to compare and select answers. The mixed dataset achieves the best performance at 59.0\%,
suggesting the two formats provide complementary supervision signals:
mcQTSA sharpens answer selection while ndQTSA strengthens reasoning
and explanatory depth. 

Accordingly, all subsequent fine-tuning
experiments adopt a mixed-format training set of 17M tokens,
comprising the full ndQTSA corpus and the non-benchmark partition
of mcQTSA.
Table~\ref{tab:Finetune_model} summarizes the results, where -T and
-NT denote thinking and non-thinking inference modes.
\vspace{-1em}
\begin{table}[htbp]
\centering
\small
\renewcommand{\arraystretch}{1.2}
\setlength{\tabcolsep}{4pt}
\caption{Fine-tuning Results on the RF Benchmark}
\label{tab:Finetune_model}
\begin{tabular}{lcccc}
\toprule
\textbf{Model} & \textbf{Base} & \textbf{Avg.} & \textbf{Fine-tuned}
               & \textbf{Avg.} \\
               & \textbf{Acc.} & \textbf{Time (s)} & \textbf{Acc.}
               & \textbf{Time (s)} \\
\midrule
\textbf{Llama3.2-1B-Base}     & 24.7\% & 1.34   & 54.4\% & 18.80 \\
\textbf{Llama3.2-1B-Instruct} & 34.2\% & 3.75   & 59.3\% & 31.10 \\
\midrule
\textbf{Llama3.2-3B-Base}     & 37.1\% & 42.64  & 70.5\% & 45.64 \\
\textbf{Llama3.2-3B-Instruct} & 71.3\% & 2.49   & 74.7\% & 16.50 \\
\midrule\midrule
\textbf{Qwen3-0.6B-T}  & 63.6\% & 62.98  & 70.2\% & 35.87 \\
\textbf{Qwen3-0.6B-NT} & 62.9\% & 4.57   & 67.7\% & 16.22 \\
\midrule
\textbf{Qwen3-1.7B-T}  & 70.3\% & 68.47  & 75.3\% & 41.61 \\
\textbf{Qwen3-1.7B-NT} & 66.5\% & 2.29   & 71.3\% & 17.94 \\
\midrule
\textbf{Qwen3-4B-T}    & 82.6\% & 112.78 & 83.5\% & 53.14 \\
\textbf{Qwen3-4B-NT}   & 78.2\% & 5.06   & 80.5\% & 32.59 \\
\bottomrule
\end{tabular}
\vspace{-0.5em}
\end{table}

\subsubsection{Domain Fine-Tuning Consistently Improves RF Reasoning}
Domain-specific fine-tuning improves model performance across all
architectures and sizes, with gains ranging from modest improvements
on already-capable models to dramatic lifts on smaller base models.
For instance, Llama3.2-3B-Base improves from 37.1\% to 70.5\%, demonstrating that
QTSA provides effective supervision for adapting general-purpose
language models to RF-domain reasoning, consistent with prior work
for the digital domain~\cite{eda_raft} and analog
domain~\cite{analogseeker}.

\subsubsection{Thinking Mode Boosts Accuracy While Fine-Tuning Cuts Latency}
Qwen3-T models consistently outperform their Qwen3-NT counterparts before and after fine-tuning. This
aligns with the Q+T+SA training format of QTSA, which supervises the
full reasoning trajectory rather than the final answer
alone~\cite{ho2023cot_teachers}. Notably, fine-tuning has opposite
effects on inference time depending on the inference mode. For NT and
base models, inference time increases after fine-tuning as the model
learns to generate structured reasoning chains rather than short or
degenerate outputs, for example, Qwen3-1.7B-NT increases from
2.29\,s to 17.94\,s. In contrast, T models become faster after
fine-tuning: Qwen3-1.7B-T from 68.47\,s to 41.61\,s. This suggests that domain
supervision focuses internal reasoning process and reduces
unnecessary exploration, yielding accuracy and efficiency gains
for T models.

\subsubsection{Instruction Tuning and Domain Fine-Tuning Are Complementary}
Instruction-tuned variants consistently exhibit higher
pre-fine-tuning accuracy than their base counterparts, confirming
that general instruction alignment helps models interpret
question--answer style inputs. After fine-tuning, this gap
narrows substantially, for example, Llama3.2-1B-Base reaches
54.4\% versus 59.3\% for the instruct variant, suggesting domain-specific supervision partially compensates for the absence
of prior instruction tuning.

\subsubsection{Fine-Tuning Gains Diminish for Larger, Stronger Models}
For larger models such as Qwen3-4B, which already achieve strong
baseline performance, fine-tuning yields only marginal gains
($<$2.5\%). Further improvement likely requires larger or more
diverse training data~\cite{llm_train}. A scaling study on
Llama3.2-1B confirms that accuracy saturates near 9M tokens,
indicating that the 17M-token QTSA provides sufficient headroom for
small models ($\leq$1B parameters) while larger models ($\geq$4B
parameters) may benefit from additional data.

\subsection{RAG Results}
Because our QA benchmark contains no diagram figures, including
diagram-derived chunks changed accuracy by $<$0.5\%. Reported results
therefore exclude them.

\subsubsection{RAG Improves SOTA Model Performance}

Table~\ref{tab:rag_sota} reports QA accuracy for three state-of-the-art
LLMs with and without our three RAG configurations. All three configurations improve accuracy over the no-RAG baseline, demonstrating that the RF
knowledge base provides domain-specific grounding that benefits even
large models. Semantic retrieval yields the strongest absolute
gains, bringing GPT-4o from 89.6\% to 93.0\%, Qwen3-235B from 87.9\% to 91.2\%, and DeepSeek-V3.2-T
from 89.1\% to 93.7\%.

\begin{table}[htbp]
\centering
\small
\renewcommand{\arraystretch}{1.1}
\setlength{\tabcolsep}{4pt}
\caption{Evaluation of State-of-the-Art LLMs With and Without RAG}
\label{tab:rag_sota}
\begin{tabular}{lcccc}
\toprule
\textbf{Model} & \textbf{No RAG} & \textbf{Semantic}
               & \textbf{Keyword} & \textbf{Hybrid} \\
\midrule
GPT-4o            & 89.6\% & 93.0\% & 92.1\% & 91.2\% \\
Qwen3-235B-A22B-T & 87.9\%   &    91.2\%   &    90.9\%   &    90.8\%   \\
DeepSeek-V3.2-T   & 89.1\% & 93.7\% & 93.0\% & 91.6\% \\
\bottomrule
\end{tabular}
\vspace{-2em}
\end{table}

\subsubsection{Retrieval Quality and Model Scaling}
To confirm the gains stem from retrieval quality rather than incidental
context injection, we run a hit-and-miss experiment on 100 questions
across three Qwen3 sizes: a \emph{hit} uses the top-3 retrieved chunks,
a \emph{miss} the next three (ranks 4--6) from the same run. As shown in
Table~\ref{tab:hit_miss}, hit chunks consistently outperform miss chunks
across all configurations and sizes, confirming retrieval quality drives
the gains. Hit accuracy under semantic RAG also rises with model size,
from 81\% (0.6B) to 84\% (1.7B) to 92\% (4B).
\vspace{-2em}

\begin{table}[htbp]
\centering
\small
\renewcommand{\arraystretch}{1.1}
\setlength{\tabcolsep}{4pt}
\caption{Hit-and-Miss Retrieval Validation Across Model Sizes}
\label{tab:hit_miss}
\begin{tabular}{lcccccc}
\toprule
\textbf{Model}
    & \multicolumn{2}{c}{\textbf{Semantic RAG}}
    & \multicolumn{2}{c}{\textbf{Keyword RAG}}
    & \multicolumn{2}{c}{\textbf{Hybrid RAG}} \\
\cmidrule(lr){2-3}\cmidrule(lr){4-5}\cmidrule(lr){6-7}
    & \textbf{Hit} & \textbf{Miss}
    & \textbf{Hit} & \textbf{Miss}
    & \textbf{Hit} & \textbf{Miss} \\
\midrule
Qwen3-0.6B-T & \textbf{81\%} & 65\% & \textbf{74\%} & 73\% & \textbf{71\%} & 65\% \\
Qwen3-1.7B-T & \textbf{84\%} & 75\% & \textbf{80\%} & 75\% & \textbf{80\%} & 75\% \\
Qwen3-4B-T   & \textbf{92\%} & 83\% & \textbf{86\%} & 82\% & \textbf{85\%} & 80\% \\
\midrule
\textit{Avg.\ $\Delta$} & \multicolumn{2}{c}{$+10.3\%$}
                        & \multicolumn{2}{c}{$+3.3\%$}
                        & \multicolumn{2}{c}{$+5.0\%$} \\
\bottomrule
\end{tabular}
\vspace{-1em}
\end{table}

\subsubsection{Comparative Analysis of Retrieval Configurations}
Tables~\ref{tab:rag_sota} and~\ref{tab:hit_miss} consistently rank
semantic retrieval highest, then keyword, then hybrid, with the
hit--miss gap corroborating the order (+10.3\%, +3.3\%, +5.0\%). Keyword
underperforms because RF reasoning favors conceptual, derivation-level
relevance that dense embeddings capture better than lexical overlap.
Hybrid underperforms because merging dense and sparse candidates before
re-ranking injects noisier context than the focused semantic top-3;
RRF~\cite{rrf} ranks by position to mitigate score mismatch, but
re-ranking cannot fully recover precision when the pool contains
low-quality BM25 candidates.

\section{Conclusion} \label{sec:conclusion}
We presented RF-Agent, a framework for LLM domain adaptation in RF
circuit design, contributing the RF-domain QTSA dataset, a standardized
multiple-choice benchmark, and a systematic study of SFT and RAG
adaptation strategies for RF reasoning.
Several directions remain open. The current benchmark may limit
discrimination among stronger models, motivating harder tasks beyond
multiple-choice QA toward agentic design scenarios. RLHF with domain
experts could provide richer supervision than distillation alone, and
expert evaluation of model-generated reasoning would validate practical
utility beyond automated benchmarks.

\vspace{-1em}
\section*{Acknowledgment}
The authors thank Xin Zhang, Luyao Shi, Prashanth Vijayaraghavan,
and Ehsan Degan of IBM Research for their insightful discussions
and feedback throughout this work.

\bibliographystyle{IEEEtran}
\bibliography{references}

\appendices

\section{A Representative mcQTSA Sample}
\label{app:mcqtsa}
\begin{figure}[t]
\centering
\includegraphics[height=12.5cm, trim=30 10 25 15, clip]{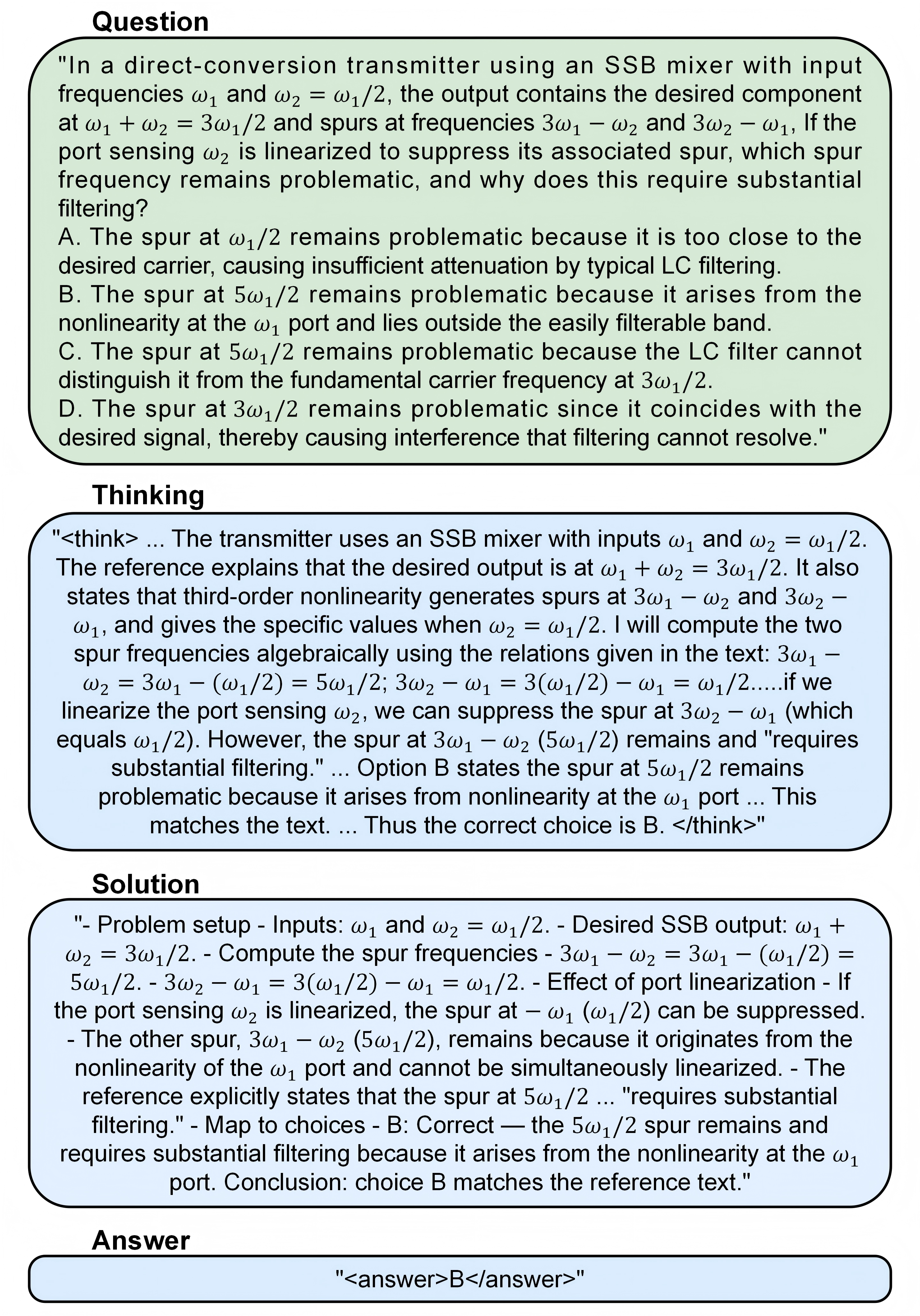}
\vspace{-1em}
\caption{A representative mcQTSA sample for mixers.}
\label{fig:qtsa_example}
\vspace{-2.5em}
\end{figure}

Fig.~\ref{fig:qtsa_example} shows a complete mcQTSA sample produced by our
QTSA pipeline, including the question, four options, and the full
Thinking--Solution--Answer trace used as SFT supervision.

\section{A Representative RAG Hit-vs-Miss Case}
\label{app:hitmiss}
\begin{figure}[htbp]
    \centering
    \includegraphics[width=8cm,clip]{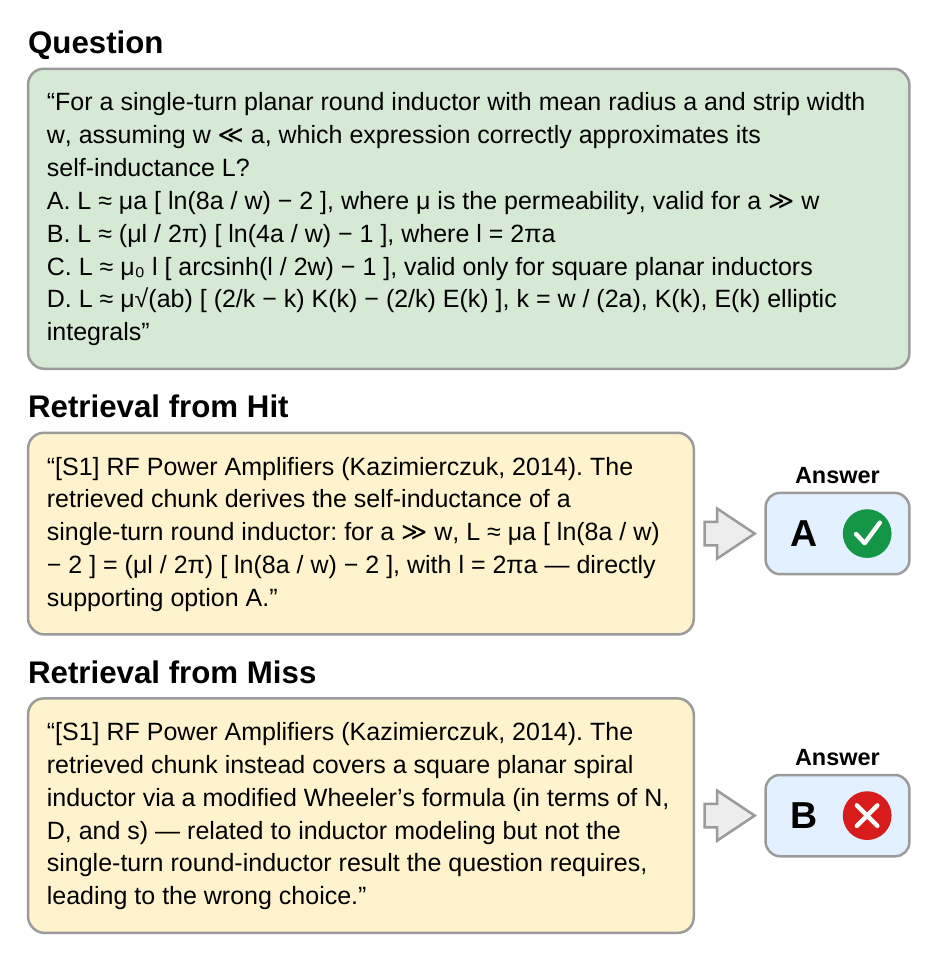}
    \vspace{-1em}
    \caption{A representative hit-and-miss example. The hit chunk
    (top) retrieves the directly relevant derivation, yielding a
    correct answer. The miss chunk (bottom) retrieves a related but
    insufficient passage from the same document, yielding an
    incorrect answer. Retrieved passages are summarized for brevity.}
    \label{fig:hit_and_miss}
\end{figure}

Fig.~\ref{fig:hit_and_miss} illustrates the hit-and-miss study. For one benchmark
question, the \emph{hit} condition uses the top-3 retrieved chunks and the
\emph{miss} condition uses ranks 4--6 from the same run: the hit chunk surfaces
the exact derivation (correct answer), while the miss chunk returns a related
but insufficient passage (incorrect answer).
\vspace{-1em}

\end{document}